\pdfoutput=1

\documentclass[11pt]{article}

\usepackage[]{acl}

\usepackage{times}
\usepackage{latexsym}

\usepackage[T1]{fontenc}

\usepackage[utf8]{inputenc}

\usepackage{microtype}

\usepackage{inconsolata}

\usepackage{graphicx}
\usepackage{enumitem}
\usepackage{multirow}
\usepackage{adjustbox}
\usepackage{booktabs}
\usepackage[ruled,vlined]{algorithm2e}
\usepackage{amsmath}
\usepackage{array}
\usepackage{pifont}
\usepackage{xcolor} 

\title{Efficient Layer-wise LLM Fine-tuning for Revision Intention Prediction}

\author{Zhexiong Liu, Diane Litman
 \\
    Department of Computer Science, 
  Learning Research \& Development Center \\
University of Pittsburgh, Pittsburgh, Pennsylvania, USA 15260
\\
   \texttt{zhexiong@cs.pitt.edu}, \texttt{dlitman@pitt.edu}
  }

\begin{document}
\maketitle
\begin{abstract}
Large Language Models (LLMs) have shown extraordinary success across various text generation tasks; however, their potential for simple yet essential text classification remains underexplored, as LLM pre-training tends to emphasize generation over classification. While LLMs with instruction tuning can transform classification into a generation task, they often struggle to categorize nuanced texts. One such example is text revision, which involves nuanced edits between pairs of texts. Although simply fine-tuning LLMs for revision classification seems plausible, it requires a large amount of revision annotations, which are exceptionally expensive and scarce in the community. To address this issue, we introduce a plug-and-play layer-wise parameter-efficient fine-tuning (PEFT) framework, i.e., IR-Tuning, which fine-tunes a subset of important LLM layers that are dynamically selected based on their gradient norm distribution, while freezing those of redundant layers. Extensive experiments suggest that IR-Tuning surpasses several layer-wise PEFT baselines over diverse text revisions, while achieving fast convergence, low GPU memory consumption, and effectiveness on small revision corpora.  
\end{abstract}

\section{Introduction}
\label{sec:introduction}

\begin{figure}[t]
  \includegraphics[width=1\columnwidth]{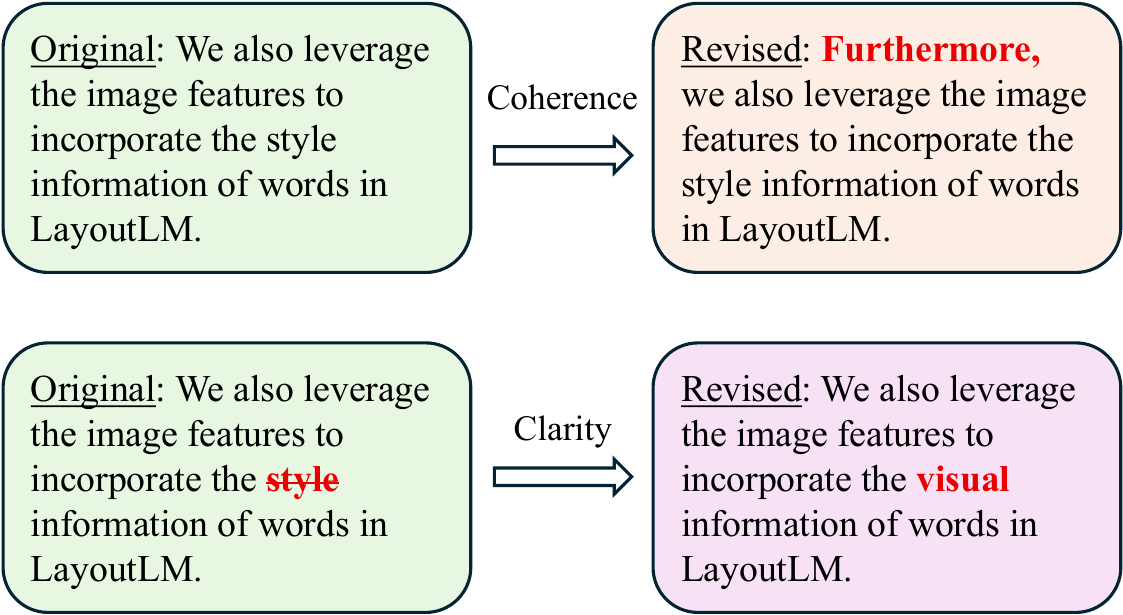}
  \caption{An example where the same original text is revised based on different intentions. The examples are from the ITERATER corpus~\cite{du-etal-2022-understanding-iterative}.}
  \label{fig:intro-example}
\end{figure}

Revision is regarded as an important part of writing because it commonly improves the final written work~\cite{fitzgerald1987research}; however, as complexity in the process of revision increases, it becomes more difficult to interpret whether revision is in line with the writer's actual intentions~\cite{sommers1980revision,hayes1986writing}. Particularly, determining the intentions between nuanced revisions is challenging for computational models. For example, Figure~\ref{fig:intro-example} shows the same piece of text revised with different intentions, of which one added a word \textit{furthermore} to improve coherence, and the other modified a word from \textit{style} to \textit{visual} to enhance clarity. Prior work~\cite{du-etal-2022-read,du-etal-2022-understanding-iterative} has used small language models, e.g., RoBERTa~\cite{liu2019robertarobustlyoptimizedbert}, to learn these intentions, which failed to identify complex revisions due to the model's limited capability~\cite{skitalinskaya-wachsmuth-2023-revise}. Hence, stronger models are needed to recognize diverse revision patterns. 

\begin{figure*}[t]
\centering
  \includegraphics[width=0.94\linewidth]{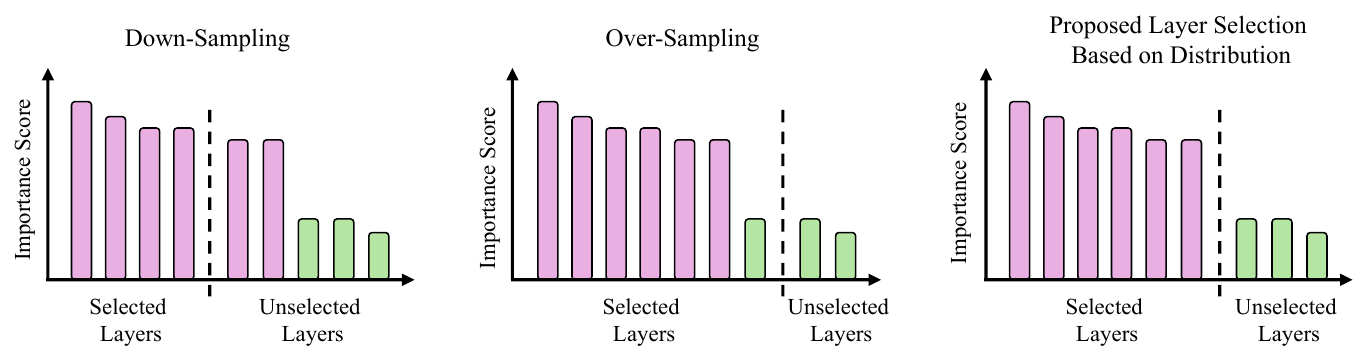}
  \vspace{-.5em}
  \caption{The illustration of over-sampling and down-sampling issues in the sampling-based layer-wise PEFT method~\cite{yao-etal-2024-layer} and our proposed layer selection method, where the purple and green bars represent high and low layer-wise importance scores introduced in Section~\ref{sec:layer_selection}, respectively.}
  \vspace{-.5em}
  \label{fig:over-down-sampling}
\end{figure*}

Recently, large language models (LLMs) have achieved impressive success in multiple NLP tasks, such as text summarization~\cite{takeshita-etal-2024-aclsum}, question answering~\cite{peng2024chain}, and concept reasoning~\cite{li2024cr}. However, their application in revision tasks remains underexplored. This might be because revision tasks require LLMs to learn an iterative process involving additions, deletions, and modifications, each driven by distinct intentions, but LLMs are mostly pre-trained to generate final texts. Although~\citet{Shu_Luo_Hoskere_Zhu_Liu_Tong_Chen_Meng_2024} explore LLMs with instruction tuning and reinforcement learning for text revision, they focus on supervised fine-tuning using massive datasets, which becomes inefficient and expensive as the data size grows. While~\citet{ruan-etal-2024-large} fine-tune LLMs using a parameter-efficient fine-tuning (PEFT) method, i.e., QLoRA~\cite{dettmers-qlora-2023}, to minimize expenses, it requires training an adapter for every LLM layer, thus becoming less efficient as the number of LLM layers increases. 

Although standard PEFT methods~\cite{hu2022lora,hu-etal-2023-llm} proved to be efficient, they attempt to learn entire LLM layers, despite prior research indicating that LLM layers contribute differently to downstream tasks~\cite{pan2024lisa,zhao-etal-2024-layer}. Recently,~\citet{yao-etal-2024-layer} have introduced an Importance-aware Sparse Tuning (IST) that samples a subset of important LLM layers for PEFT. However, the number of its important layers remains fixed throughout the fine-tuning, leading to suboptimal layer selection. For example, important layers might be overlooked (down-sampling) if sampling a small number of layers, or unimportant layers might be included (over-sampling) if selecting too many layers. Figure~\ref{fig:over-down-sampling} illustrates these scenarios, where down-sampling selects four out of six important layers and over-sampling selects an extra unimportant layer. Also, the number of important layers can vary as the fine-tuning progresses, but IST failed to address it. In contrast, we propose Importance Redundancy Tuning, i.e., IR-Tuning\footnote{\url{https://github.com/ZhexiongLiu/IR-Tuning}}, which uses an algorithm to select important layers for fine-tuning while freezing redundant ones, where the fine-tuned layers are dynamically determined based on the distribution of LLM layer-wise gradient norms throughout the fine-tuning.

To evaluate the effectiveness of the IR-Tuning on text revision tasks, we work on three research questions:
\textbf{RQ1}: How do LLM layers contribute differently to revision intention prediction?
\textbf{RQ2}: How can we dynamically select important layers for fine-tuning LLMs on small corpora?
\textbf{RQ3}: Are contextualized instructions helpful for LLMs to learn revision intentions? In particular, we make the following contributions:

\begin{itemize}
[leftmargin=*,itemsep=-1pt, topsep=3pt]
\item We propose the first work that uses layer-wise PEFT to address an essential text revision task.
\item We develop an algorithm to dynamically select a subset of important LLM layers for fine-tuning.
\item We demonstrate the feasibility of efficiently fine-tuning LLMs with small revision corpora.

\end{itemize}

\section{Related Work}
\subsection{Revision Intention in NLP} 
Text revision primarily focuses on analyzing revision intention to understand human edits~\cite{zhang2015l,shibani2018kb,afrin2020RER,kashefi2022argrewrite,du-etal-2022-understanding-iterative,chong-etal-2023-leveraging,mita-etal-2024-towards,jourdan-etal-2024-casimir}. However, identifying intention is challenging, largely because collecting annotated revision corpora are expensive~\cite{zhang-etal-2017-corpus,anthonio-etal-2020-wikihowtoimprove,spangher-etal-2022-newsedits,du-etal-2022-understanding-iterative,darcy-etal-2024-aries}. Prior work~\cite{argrewrite, wikieditintent-yang, afrin2018improvement, kashefi2022argrewrite,afrin2020RER,afrin2023predicting} has developed feature-based computational methods, which cannot generalize to other corpora. Although~\citet{du-etal-2022-understanding-iterative, arxivedits} have trained small language models, such as RoBERTa~\cite{liu2019robertarobustlyoptimizedbert}, for identifying revision intention, they have struggled to learn complex revision patterns due to the models' capabilities. While more recent work either prompts LLMs with instruction~\cite{ruan-etal-2024-re3} or fine-tunes LLMs using standard PEFT~\cite{ruan-etal-2024-large}, they have not investigated how LLM layers contribute to revision tasks. In contrast, we propose a novel layer-wise PEFT method that facilitates LLM fine-tuning more effectively and efficiently using small revision corpora. 

\begin{table*}[t]\small
\centering
\begin{adjustbox}{width=1.6\columnwidth}
\begin{tabular}{c|cccc|cccc|c}
\toprule
\multirow{2}{*}{} & \multicolumn{4}{c|}{Evidence Revision}    & \multicolumn{4}{c|}{Reasoning Revision} & \multirow{3}{*}{Total} \\ \cmidrule{2-9}
                  & Relevant & Irrelevant & Repeated & Others & LCE    & not LCE  & Commentary & Others &                        \\ \midrule
Add               & 1,774    & 397        & 225      & 136    & 1,069  & 317      & 425        & 299    & 4,642                  \\
Delete            & 598      & 102        & 45       & 56     & 318    & 121      & 194        & 70     & 1,504                  \\
Modify            & 138      & 26         & 6        & 53     & 105    & 23       & 41         & 55     & 447                    \\ \midrule
Total             & 2,510    & 525        & 276      & 245    & 1,492  & 461      & 660        & 424    & 6,593                  \\ \bottomrule
\end{tabular}
\end{adjustbox}
\caption{
The statistics of revisions in the ArgRevision corpus.
}
\label{table:sentence-stats}
\vspace{-1em}
\end{table*}

\subsection{Parameter-Efficient Fine-tuning}
PEFT methods offer promising solutions for fine-tuning LLMs in a computationally efficient manner. They update a small fraction of LLM parameters by using adapter-based techniques~\cite{pmlr-v97-houlsby19a,wang2022adamix,lei2024conditional}, low-rank adaptations~\cite {hu2022lora,krona,liu-etal-dora-2024}, and prompt-based approaches~\cite{prompttuning,li-liang-2021-prefix,ptuning}. While standard PEFT implements an identical design across all LLM layers, it cannot explore each layer's unique contribution to downstream tasks, despite prior work suggesting layer redundancy in pre-trained models~\cite{layersharing,layerdrop,crash,elhoushi2024layer}. Recently,~\citet{kaplun2023less} develop a greedy search to select useful layers,~\citet{pan2024lisa} instead randomly select layers, and~\citet{zhu2023lift} pick layers based on front-to-end or end-to-front heuristics, all for fine-tuning LLMs, but these methods either need expensive computations or utilize simple strategies that could cause downgraded performance on complex tasks. While prior work~\cite{yao-etal-2024-layer,zhou-etal-2025-lora} sampled a subset of important layers based on importance scores, their methods could cause over-sampling and down-sampling issues, as shown in Figure~\ref{fig:over-down-sampling}. Although~\citet{wei-etal-2025-flexora} utilize an unrolled differentiation method to identify the most useful LLM layers, they require expensive computation on hyperparameter optimization. In contrast, we propose a plug-and-play PEFT framework that utilizes an efficient algorithm to dynamically select a subset of important LLM layers based on their gradient norm distribution in each fine-tuning iteration, ensuring high-gradient layers are prioritized for update.

\section{Corpora}
Text revision is rarely annotated because annotation is expensive; thus, we collect an argument revision corpus called ArgRevision, which includes essays written by elementary and middle school students with limited writing skills. We use this corpus to study revisions in argumentation, which is critically needed for argument writing evaluation research~\cite{li-etal-2024-using,correnti2024supporting}. Additionally, we use a publicly available revision corpus~\cite{du-etal-2022-understanding-iterative} that contains articles written by experienced authors, editors, and researchers. We conduct experiments using text revisions from both skilled and less skilled writers. 

\begin{table}[t]\small
\centering
\begin{adjustbox}{width=0.75\columnwidth}
\begin{tabular}{c|cccc}
\toprule
          & \multicolumn{2}{c}{{Space Essays}} & \multicolumn{2}{c}{{MVP Essays}} \\ \cmidrule{2-5} 
          &  RER\#            & Kappa            & RER\#            & Kappa           \\ \midrule
Reasoning & 148                 & 0.86             & 135                 & 0.84            \\
Evidence  & 108                 & 0.89             & 136                 & 0.80            \\ \bottomrule
\end{tabular}
\end{adjustbox}
\caption{The annotation agreement for reasoning and evidence RER annotations for a batch of 117 essays in our prior studies~\cite{liu-etal-2023-predicting}.}
\vspace{-1em}
\label{table:annotation-agreement}
\end{table}

\subsection{Argument Revision}
The ArgRevision corpus contains pairs of essays before and after revisions, collected using our deployed automated writing evaluation system~\cite{liu2025ereviserfwritingevaluationassessing}. The corpus contains 990 essay drafts written by students in grades four to eight from schools in Pennsylvania and Louisiana, who are taking the Response to Text Assessment~\cite{correnti2013assessing}.  172 students wrote essays in response to a prompt about the United Nations' Millennium Villages Project (MVP). Afterward, the students revised their essay drafts in response to feedback provided by the system, and each student completed three drafts, resulting in 344 pairs of essay drafts, e.g., draft1-draft2 and draft2-draft3. Another 158 students did the same tasks in response to another essay prompt about Space Exploration (Space), yielding 316 pairs of essay drafts. We combine the essays from the two prompts as students share similar writing skills, and the scoring rubric is consistent across the prompts.

\begin{figure*}[t]
    \centering
    \includegraphics[width=1\linewidth]{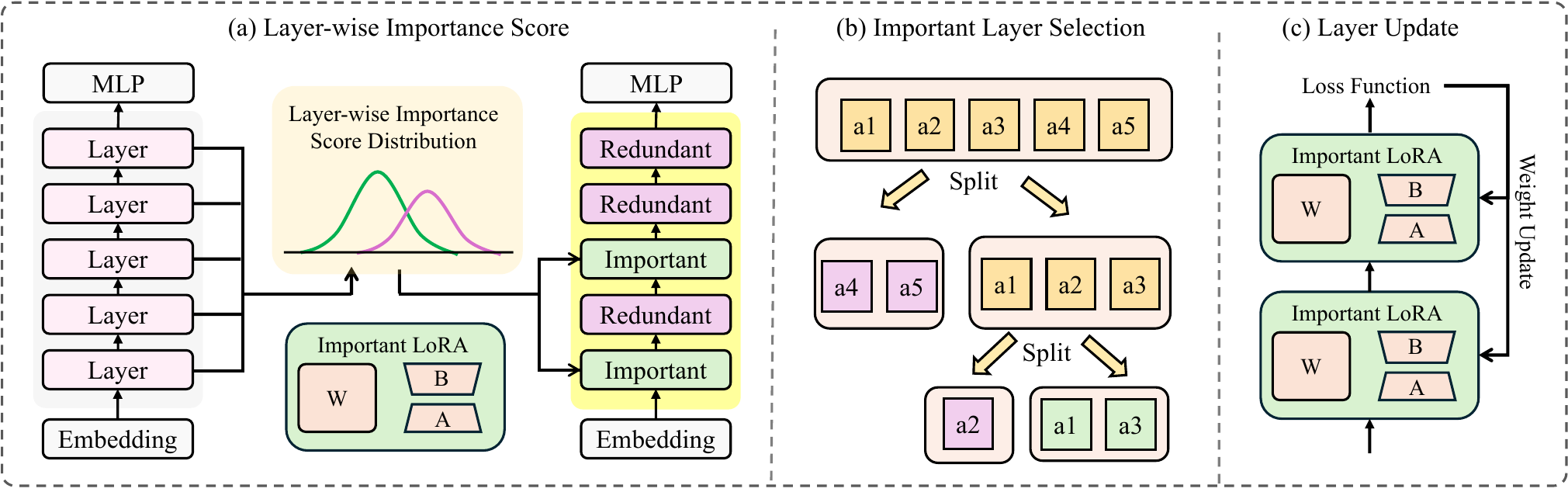}
    \vspace{-1.8em}
    \caption{The components of the proposed Importance Redundancy Tuning (IR-Tuning). The IR-Tuning includes a module that updates layer-wise importance scores on all the layers, then a layer-selection module to select important layers for fine-tuning. Finally, the weights in the selected layers are updated with fine-tuning through LoRA.}
    \vspace{-0.5em}
    \label{fig:ir-tuning}
\end{figure*}

We preprocess collected essays for revision annotation. First, the sentences from the original and revised drafts (e.g., draft1-draft2, draft2-draft3) are aligned into pairs of original sentences and revised sentences using a sentence alignment tool Bertalign~\cite{liu2022bertalign}. The aligned pairs are automatically labeled with \textit{no change} if the original sentence and the revised sentence are the same, \textit{modify} if the original and the revised sentence are not empty but not the same, \textit{add} if the original sentence is empty but the revised sentence is not, or \textit{delete} if the revised sentence is empty but the original sentence is not. The changed alignments are classified into \textit{surface} (meaning-preserving) and \textit{content} (meaning-altering) revisions by a BERT-based~\cite{devlin-etal-2019-bert} classifier trained on a college-level revision corpus~\cite{kashefi2022argrewrite} with an F1 score of 0.96. Following~\citet{afrin2020RER}, we use the Revisions of Evidence and Reasoning (RER) scheme to annotate \textit{content} revisions into evidence and reasoning revisions. Specifically, evidence revisions are annotated with \textit{relevant, irrelevant, repeated} evidence and \textit{others}, and reasoning revisions are annotated with \textit{linked claim-evidence (LCE), not LCE, commentary} and \textit{others}. Here, we do not use the~\textit{others} label as it contains a mixture of revisions based on multiple rarely annotated intentions. Table \ref{table:sentence-stats} shows annotated intention statistics. The annotation is done by expert annotators, and each revision is annotated by one expert. Some annotators participated in the same annotation tasks in our prior studies~\cite{liu-etal-2023-predicting}, of which the two-annotator agreements on a batch of 117 student essays about both MVP and Space prompts are shown in Table~\ref{table:annotation-agreement}. The annotation examples are shown in Table~\ref{table:data-example-argrevision} in the Appendix. 

\subsection{Article Revision}
We use a publicly available revision corpus named ITERATER~\cite{du-etal-2022-understanding-iterative}, which annotates 4,018 text revisions from Wikipedia, ArXiv, and news articles.  Wikipedia articles are written by editors who often focus on improving the clarity and structure of articles. ArXiv articles are written by scientific authors who generally revise hypotheses, experimental results, and research insights. News articles are written by editors interested in clarity and readability. The ITERATER corpus contains six intention labels:  \textit{clarity}, \textit{fluency}, \textit{coherence}, \textit{style}, \textit{meaning-changed}, and \textit{others}. Here, we do not use the \textit{others} label as it denotes unrecognizable intentions. The revision statistics and examples are shown in Table~\ref{table:iterator-corpus-stats1} and Table~\ref{table:data-example-iterater} in the Appendix. 

\section{Method}
\label{sec:layer-select-overview}
We propose a novel plug-and-play layer-wise PEFT framework, i.e., IR-Tuning, which attempts to update the weights of important LLM layers while keeping those of redundant layers unchanged. Figure~\ref{fig:ir-tuning} shows the framework components. 

\subsection{Layer-wise Importance Score}
Supposing an LLM $\mathcal{M}$ consisting of $l$ transformer layers $m_i$, $\mathcal{M}=\left\{m_i\right\}_{i=1}^{l}$, our objective is to generate a subset $\mathcal{S}$ of $\mathcal{M}$ as important layers and the remaining set $\bar{\mathcal{S}}$ as redundant layers. Inspired by prior work~\cite{Zhang_2024_CVPR}, we use layer-wise gradient norm $\mathcal{N}=\{a_i\}_{i=1}^{l}$ to denote LLM layers' importance scores for two reasons. First, layers with high gradient norms suggest large weight updates that make key contributions to the rapid optimization of the loss function along the gradient direction, which can facilitate efficient gradient descent. Second, larger gradient norms may carry more information relevant to downstream tasks, which makes LLM layers informative during fine-tuning. These hypotheses have been confirmed and utilized in prior work~\cite{lee2023surgical,Zhang_2024_CVPR}. Empirically, we use a threshold $\gamma$ to split $\mathcal{M}$ into $\mathcal{S}$ and $\bar{\mathcal{S}}$:
\begin{equation}
\mathcal{S}=\left\{m_i \mid a_i>\gamma\right\},
\label{eq:e1}
\end{equation}
\begin{equation}
    \bar{\mathcal{S}}=\left\{m_i \mid a_i \leq \gamma\right\},
    \label{eq:e2}
\end{equation}
\noindent where $\gamma$ is obtained by a layer selection algorithm described in the next section.

\subsection{Important Layer Selection}
\label{sec:layer_selection}
Unlike prior work that selects the top $n$ important layers from a ranked score list~\cite{yao-etal-2024-layer}, we formulate the layer selection as a distribution divergence problem, arguing that the important layers and redundant layers are from different distributions. Although this idea has been used in~\citet{Liu_Liu_Xie_Jin_Jia_2023} for splitting a large amount of meta-learning tasks into easy and hard tasks, we instead focus on a smaller set of LLM layer splitting based on their importance scores. Here, we have a null hypothesis $\mathcal{H}_0$: the importance scores of layers in $\mathcal{M}$ follows a single distribution, and an alternative hypothesis $\mathcal{H}_1$: there exists a subset $\mathcal{S}$ of $\mathcal{M}$, where the importance scores of layers in $\mathcal{S}$ follow a different distribution from those of the remaining layers $\bar{\mathcal{S}}$. Therefore, the optimal set $\mathcal{S}^*$ can be obtained by solving the optimization problem:
\begin{equation}
    \mathcal{S}^*=\operatorname{argmax}_\mathcal{N} \log \frac{\text { Likelihood }\left(\mathcal{H}_1 \mid \mathcal{N}\right)}{\text { Likelihood }\left(\mathcal{H}_0\right)}.
\end{equation}
\noindent Intuitively, we need to maximize the likelihood of the alternative hypothesis that suggests the layer-wise importance scores in $\mathcal{S}$ and $\bar{\mathcal{S}}$ are different, where the former is denoted as $\mathcal{N}_{\mathcal{S}}=\left\{{a_i\mid m_i\in\mathcal{S}}\right\}$ and the latter is denoted as $\mathcal{N}_{\bar{\mathcal{S}}}=\left\{{a_i \mid m_i\in\bar{\mathcal{S}}}\right\}$. This can be solved by minimizing the variance of the importance scores in $\mathcal{N}_\mathcal{S}$ and $\mathcal{N}_{\bar{\mathcal{S}}}$~\cite{xie2021statistically}. Hence we choose an optimal threshold $\gamma^*$ that leads to the minimum sum of $\operatorname{Var}(\mathcal{N}_\mathcal{S})$ and $\operatorname{Var}\left(\mathcal{N}_{\bar{\mathcal{S}}}\right)$ such that
\begin{equation}
       \operatorname{Var}(\mathcal{N}_{\mathcal{S}}) + \operatorname{Var}\left(\mathcal{N}_{\bar{\mathcal{S}}}\right) <= \operatorname{Var}(\mathcal{N}).
\end{equation}

Algorithm~\ref{algo:gamma} summarizes the layer-splitting process, where the input is the full-layer importance score $\mathcal{N}$ and the output is the optimized threshold $\gamma*$, which can be used to split $\mathcal{M}$ into $\mathcal{S}$ and $\mathcal{\bar{S}}$ using Equations~\ref{eq:e1} and~\ref{eq:e2}. The algorithm is efficient, as it has a time complexity of $O(l\log l)$ with respect to LLM $\mathcal{M}$, which has $l$ layers, and is independent of the size of fine-tuning data. If we run the algorithm multiple times, we can further split the important layers $\mathcal{S}$ into more important and less important subsets. For example, a hierarchical splitting of layer-wise importance score $\mathcal{N}=\left\{a_1,a_2,a_3,a_4,a_5\right\}$ is visualized in Figure~\ref{fig:ir-tuning} (b), which yields multiple subsets, whose importance are ranked $\left\{a_1,a_3\right\}$ > $\left\{a_2\right\}$ > $\left\{a_4,a_5\right\}$ if the maximum splitting number is two. Hence, we have a fine-grained set of importance layers $\mathcal{S^*}=\left\{m_1,m_3\right\}$ used for fine-tuning if we select the highest importance score set $\left\{ a_1, a_3\right\}$. While more hierarchical splits yield more fine-grained important layers, there is a trade-off between model performance and fine-tuning efficiency, as too many splits could cause down-sampling important layers, as described in Figure~\ref{fig:over-down-sampling}.

\begin{algorithm}[t]
\SetAlgoLined
\DontPrintSemicolon
\SetKwInput{KwInput}{Input}
\SetKwInput{KwOutput}{Output}
\SetKwInput{KwLet}{Let}
\KwOutput{optimized threshold $\gamma^*$}
\KwInput{importance score array $\mathcal{N}$}
 Initialization: rank $\mathcal{N}$ in descending order; set array $\mathcal{V}=\emptyset$, $l$ the length of $\mathcal{N}$, $j=0$
 \;
 \While{$j$ is less than $l$}{
    $\mathcal{N}_{\mathcal{S}}$ $\leftarrow$ $\mathcal{N}$[$:j$] \;
    $\mathcal{N}_{\bar{\mathcal{S}}}$ $\leftarrow$ $\mathcal{N}$[$j:$] \;
    $\mathcal{V}_j$ $\leftarrow$ Var($\mathcal{N}_{\mathcal{S}}$) + Var($\mathcal{N}_{\bar{\mathcal{S}}}$) \;
    $j \leftarrow j + 1$\;
 }
 $\gamma^*$ = $\mathcal{N}$[ArgMin($\mathcal{V}$)]
 \caption{Layer splitting algorithm}
 \label{algo:gamma}
\end{algorithm}

\subsection{Layer Update}
Upon selection, the layers in $\mathcal{S}$  are encapsulated with LoRA~\cite{hu2022lora} for fine-tuning, while the layers in $\bar{\mathcal{S}}$ are frozen. Since the importance scores $\mathcal{N}$ and splitting threshold $\gamma$ keep updating as the fine-tuning progress, the sizes of $\mathcal{S}$ and $\bar{\mathcal{S}}$ might change, where important layers could become redundant and contribute less to the fine-tuning if their gradient norms descend below the optimized threshold $\gamma^*$, and vice versa. In practice, we select important and redundant layers every $k$ step(s).

\begin{figure*}[t]
    \centering
    \includegraphics[width=1\linewidth]{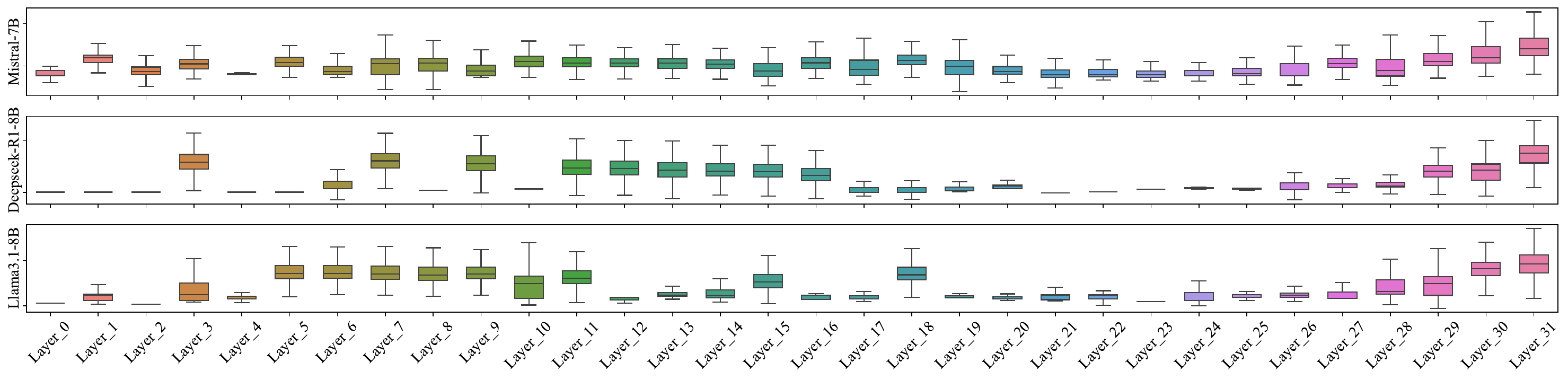}
    \vspace{-2em}
    \caption{The importance scores (gradient norms) vary across different LLM layers on the ArgRevision corpus. The high variances indicate that the gradients have changed significantly, suggesting the layers are actively learning from the data. The layers with low variances suggest they have been frozen most of the time.}
    \label{fig:gradient-change}
\end{figure*}

\section{Experiments}
\subsection{Data Preprocessing}
ArgRevision corpus contains substantial
revisions, which may have empty original sentences $\mathcal{R}_1$ (in adding revisions) or empty revised sentences $\mathcal{R}_2$ (in deleting revisions) as shown in Table~\ref{table:data-example-argrevision} in the Appendix. Revisions in ITERATER are minor, of which $\mathcal{R}_1$ and $\mathcal{R}_2$ are not null as shown in Table~\ref{table:data-example-iterater} in the Appendix. Also, ArgRevision only allows one intention for a revision, but ITERATER allows multiple, e.g., the same $\mathcal{R}_1$ could be revised to several $\mathcal{R}_2$ with different intention labels as shown in Figure~\ref{fig:intro-example} and more in rows 1, 2, and 3 in Table~\ref{table:data-example-iterater} in the Appendix. Additionally, we formulate revisions into \textbf{vanilla} and \textbf{instruction} data for vanilla fine-tuning and instruction tuning, respectively, where the former is a pair of $\left\{\mathcal{R}_1,\mathcal{R}_2\right\}$, and the latter are formulated with an instruction: \#\#\# \textit{Instruction: Identify the intention of the revision between the original sentence and the revised sentence. The possible intentions include}: $\mathcal{Y}$. \#\#\# \textit{Original Sentence}: {$\mathcal{R}_1$}. \#\#\# \textit{Revised Sentence}: {$\mathcal{R}_2$}, where $\mathcal{Y}$ is the annotated intention labels.  

\subsection{Baselines and Evaluation Metrics}
To evaluate the proposed method and answer our research questions, we use three LLMs, i.e., Mistral-7B, which is known for efficient inference~\cite{jiang2023mistral7b}, 
Llama3.1-8B, which achieves good performance for general NLP tasks~\cite{grattafiori2024llama3herdmodels}, and Deepseek-R1-8B, which is a distilled version of the Llama model with emphasized reasoning capability~\cite{guo2025deepseek}. 
We compare our IR-Tuning to the following baselines:

\begin{itemize}[leftmargin=*,itemsep=-2pt, topsep=5pt]
    \item \textbf{RoBERTa}: We use a RoBERTa-large ~\cite{devlin-etal-2019-bert} as a small language model baseline the same as~\citet{du-etal-2022-understanding-iterative}.
    \item \textbf{LISA-Baseline}: We fine-tune randomly selected four layers based on the algorithm in~\citet{pan2024lisa}. Here, we apply LoRA to the selected layers for PEFT based on~\citet{yao-etal-2024-layer}.
    \item \textbf{IST-Baseline}: We compute layer-wise importance scores based on~\citet{yao-etal-2024-layer} and sample the top eight layers for PEFT with LoRA.
    \item \textbf{Full-Finetuning}: We fine-tune full LLM layers with LoRA~\cite{hu2022lora} as an upper bound.
\end{itemize}

In the implementation, we build the framework with PyTorch\footnote{\url{https://pytorch.org}}, use pretrained models from Huggingface\footnote{\url{https://huggingface.co}}, and optimize multi-class cross-entropy loss with Adam optimizer on an Nvidia A100 GPU. We set the batch size as 16, the maximum text length cutoff as 256 for vanilla data and 1024 for instruction data, and the learning rate as 2e-4. We use a default one split to select important and redundant layers every fine-tuning step, and log results every 10 steps. We fine-tune LLMs for four epochs, and train the RoBERTa model for 20 epochs. In addition, we split the ArgRevision corpus into 80\%, 10\%, and 10\% for training, validation, and test sets, respectively. We use the splits in~\citet{du-etal-2022-understanding-iterative} for the ITERATER corpus. The detailed splits are shown in  Tables~\ref{table:sentence-stats-split} and~\ref{table:iterator-corpus-stats-split} in the Appendix. We tune hyperparameters on the validation sets and report macro-average Precision, Recall, F1-score, and the area under the precision-recall curve (AUPRC) used for evaluating imbalanced multi-class classification, all on the test sets.

\begin{table*}[t]\small
\centering
\begin{adjustbox}{width=0.99\linewidth}
\begin{tabular}{ccc|cccc|cccc}
\toprule
\multirow{2}{*}{Models}      & \multirow{2}{*}{Methods} &\multirow{2}{*}{Layer Num} & \multicolumn{4}{c}{ArgRevision Corpus}                        & \multicolumn{4}{c}{ITERATER Corpus}                               \\ \cmidrule{4-11} 
                             &            &             & Precision      & Recall         & F1-Score       & AUPRC          & Precision      & Recall         & F1-Score       & AUPRC          \\ \midrule
RoBERTa                    & -       & -         & 52.00          & 46.35          & 47.95          & 51.47          & 44.99          & 51.19          & 46.31          & 52.69          \\ \midrule
\multirow{5}{*}{Mistral-7B}  & Full-Finetuning     & fixed (32)                 & 53.60          & 50.09          & 50.54          & 49.08          & 54.95          & 52.48          & 51.56          & 56.81          \\ \cmidrule{2-11}

                             & LISA-Baseline  & fixed (4)  & {31.00}          & 34.29          & 31.06          & 43.98          & 45.31          & 47.32          & 45.14          & 51.24          \\
                             & IST-Baseline  & fixed (8)  & {44.51}          & 40.64          & 40.69          & 46.39          & 45.47          & 50.67          & 46.61          & 50.60          \\
                             & IR-Tuning (ours)  & dynamic   & \textbf{51.45} & \textbf{46.26} & \textbf{47.20} & {\,\,\,\textbf{49.16}*} & \textbf{47.61} & {\,\,\,\textbf{52.64}*} & \textbf{49.38} & \textbf{54.22} \\ \midrule

\multirow{5}{*}{DeepSeek-R1-8B} & Full-Finetuning      & fixed (32)                & 54.13          & 47.94          & 49.33          & 50.78          & 52.04          & 53.03          & 51.62          & 56.00          \\ \cmidrule{2-11}

                             & LISA-Baseline    &  fixed (4) &{\,\,\,\textbf{54.82}*}          & 38.94          & 37.48          & 44.85          & 46.94          & 50.18          & 47.26          & 51.74          \\
                             & IST-Baseline  & fixed (8) & {44.98}          & 44.27          & 43.76          & 48.64          & 47.40          & \textbf{52.11} & 48.64          & 52.76          \\
                             & IR-Tuning (ours) & dynamic   & 52.20 & \textbf{47.19} & \textbf{48.47} & \textbf{50.14} & {\,\,\,\textbf{52.93}*} & 52.06          & \textbf{50.66} & \textbf{54.46} \\ \midrule
                             
\multirow{5}{*}{Llama3.1-8B} & Full-Finetuning     & fixed (32)                 & 54.06          & 49.30          & 50.56          & 51.14          & 53.33          & 54.51          & 52.38          & 57.46          \\ \cmidrule{2-11} 
                             
                             & LISA-Baseline & fixed (4)   & 35.23          & 35.66          & 33.46          & 42.01          & 49.96          & 51.87          & 48.89          & 54.96          \\
                             & IST-Baseline  & fixed (8)  & 44.29          & 42.52          & 42.16          & 47.99          & \textbf{51.84} & \textbf{52.77} & \textbf{51.69} & 55.81          \\
                             &  IR-Tuning (ours) & dynamic   & {\,\,\,\textbf{54.17}*} & \textbf{49.17} & \textbf{50.15} & {\,\,\,\textbf{52.69}*} & 48.45          & 51.94          & 49.20          & \textbf{56.42} \\

                             \bottomrule
\end{tabular}
\end{adjustbox}
\caption{The performance of different LLMs and PEFT methods on the ArgRevision and ITERATER test sets. The bold numbers represent the best results, and the asterisks indicate that the results are better than full fine-tuning.
}
\vspace{-1em}
\label{tab:peft-comparision}
\end{table*}

\section{Results}
\subsection{The Importance of LLM Layers}
To understand the layer-wise contributions to the revision prediction, we visualize importance score (gradient norm) changes across all the LLM layers during IR-Tuning on the ArgRevision corpus, as shown in Figure~\ref{fig:gradient-change}. For Mistral-7B, gradient updates across almost all layers have relatively high variances. This suggests that all the LLM layers are actively updating and share similar capacities for handling text revisions. Regarding Deepseek-R1-8B, several layers, such as Layers 0 to 5 except Layer 3 and Layers 21 to 25, exhibit low gradient variances, which suggests these layers are not heavily involved in the fine-tuning process. Instead, layers with relatively high gradient variances, such as Layers 3, 7, 9, and several middle and top layers (Layers 29 to 31), suggest frequent updates and active learning from the data. For Llama3.1-8B, the top and bottom layers are almost identical to those of Deepseek-R1-8B, possibly because Deepseek-R1-8B was distilled on Llama models, thus sharing similar patterns. Despite differences in layer-wise gradient norms across all the LLMs, the proposed IR-Tuning can select high-gradient layers to ensure the most informative layers are effectively fine-tuned on the revision task; in contrast, less informative layers are not frequently used. Similar observations on the ITERATER corpus are shown in Figure~\ref{fig:gradient-change-iterater} in the Appendix, which confirms \textbf{RQ1} that LLM layers contribute differently to revision intention prediction, thus fine-tuning layers that contribute more while freezing those that contribute less can potentially benefit the revision task.

\subsection{The Performance of IR-Tuning} 
We show the intention prediction performance on the ArgRevision and ITERATER corpora in Table~\ref{tab:peft-comparision}. IR-Tuning achieves the best results across almost all metrics and outperforms RoBERTa in most settings, suggesting the generalizability of IR-Tuning across different LLMs and revision corpora. Although Llama3.1-8B achieves better F1 scores using IST-Baseline instead of IR-Tuning on the ITERATER corpus, the results are reversed in terms of AUPRC. This is because unbalanced data can cause the model to poorly learn the minority classes (e.g., \textit{Style} in Table~\ref{table:iterator-corpus-stats1} in the Appendix), which can downgrade the macro F1 scores that treat each class equally but are less sensitive to AUPRC. Thus, the higher AUPRC in IR-Tuning across all settings suggests its effectiveness compared to the baselines. Sometimes, LLMs with IR-Tuning can outperform full fine-tuning results on several metrics, which demonstrates the success of IR-Tuning over standard PEFT methods. These results answer \textbf{RQ2}, which suggests that we can dynamically select important layers for PEFT on even small corpora.

\begin{figure}[t]
    \centering
    \includegraphics[width=1\columnwidth]{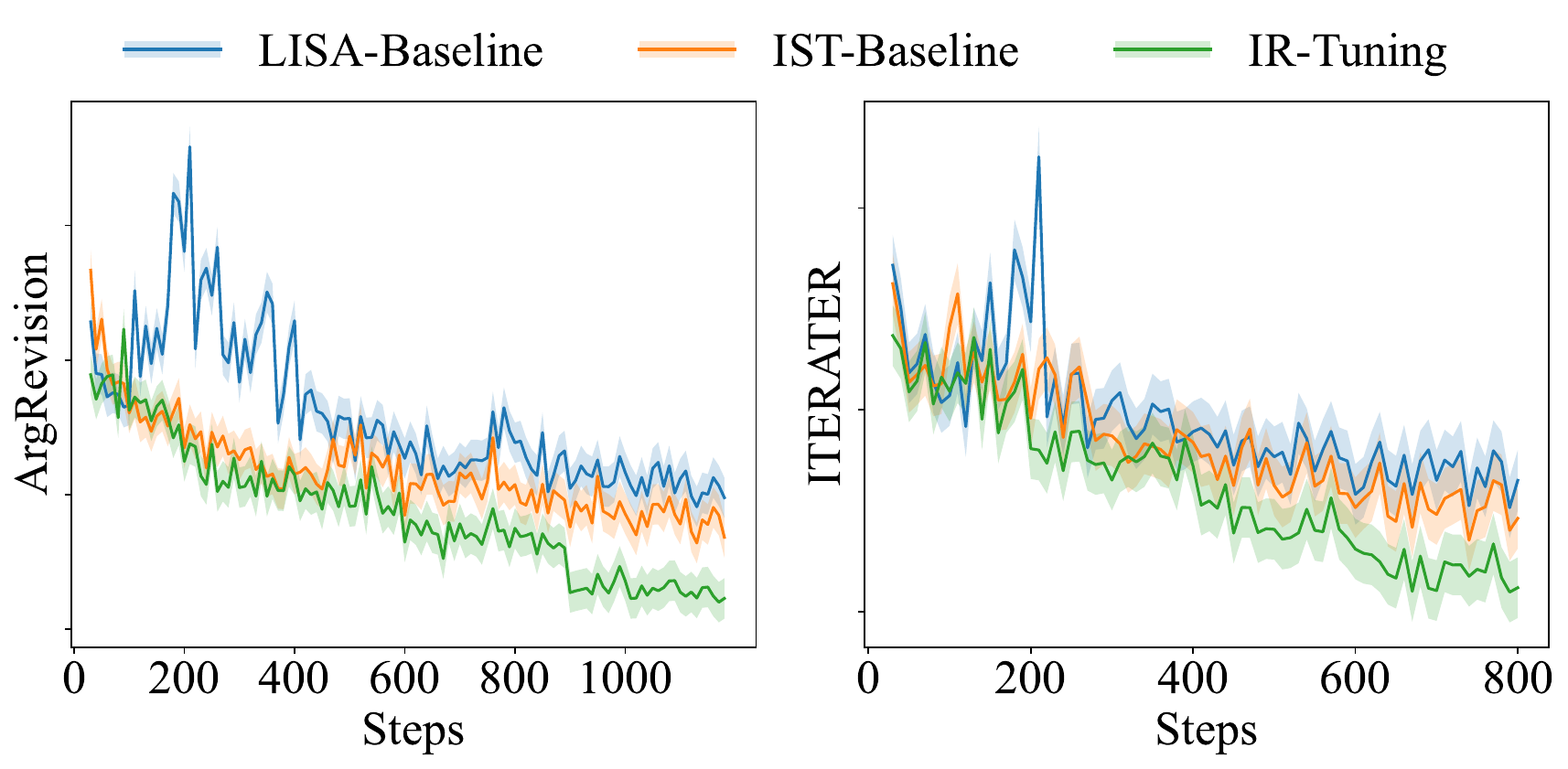}
    \vspace{-1.5em}
    \caption{The Llama3.1-8B fine-tuning losses on the ArgRevision and ITERATER training sets, using IR-Tuning and baseline PEFT methods.}
    \vspace{-0.3em}
    \label{fig:training-loss-llama3.1-8b}
\end{figure}

\begin{figure}[t]
    \centering
    \includegraphics[width=0.98\columnwidth]{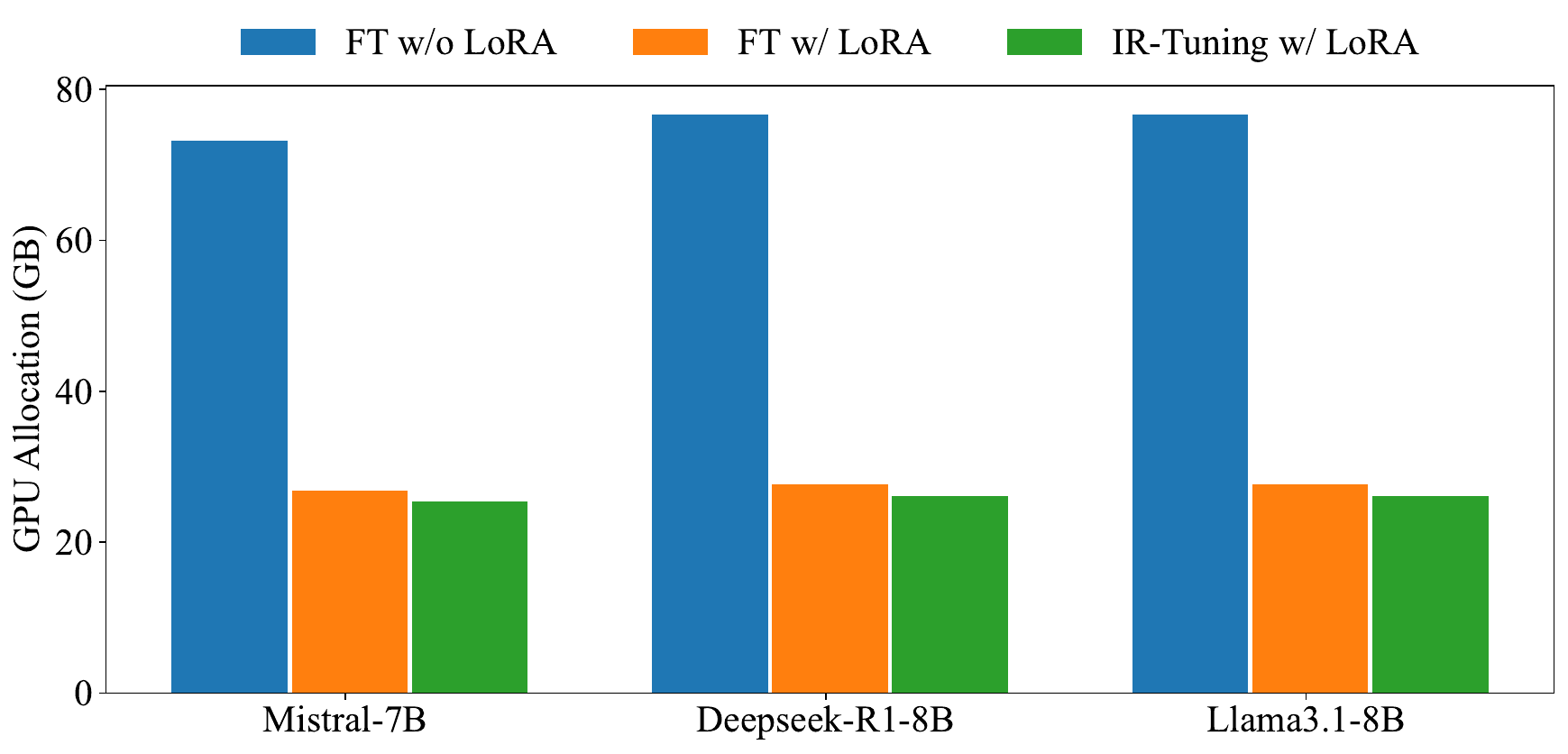}
    \caption{The GPU memory allocations for the Full-Finetuning (FT) with and without PEFT (LoRA), and IR-Tuning with PEFT (LoRA) on the ArgRevision corpus.}
    \vspace{-1em}
    \label{fig:memory-cost}
\end{figure}

\begin{table*}[tb]\small
\centering
\begin{adjustbox}{width=1\textwidth}
\begin{tabular}{c|c|ccc|ccc|ccccc}
\toprule
\multirow{2}{*}{Models} & \multirow{2}{*}{Inputs} & \multicolumn{3}{c|}{ArgRevision Evidence} & \multicolumn{3}{c|}{ArgRevision Reasoning} & \multicolumn{5}{c}{ITERATER Corpus} \\ \cmidrule{3-13} 
 &  & Relevant & Irrelevant & Repeated & LCE & not LCE& Commentary & Clarity & Fluency & Coherence & Style & Meaning \\ \midrule
RoBERTa & vanilla & 78.14 & 42.35 & 13.64 & 66.67 & 25.24 & 61.67 & 66.47 & 81.22 & 30.30 & 11.11 & 54.21 \\ \midrule
\multirow{2}{*}{Mistral-7B} & vanilla & 78.11 & \textbf{35.96} & \textbf{16.00} & \textbf{71.52} & \textbf{42.11} & \textbf{63.72} & 71.65 & 75.95 & 31.43 & 0.00 & 55.17 \\
 & instruct & \textbf{78.52} & 31.46 & 5.00 & 67.52 & 38.33 & 62.39 & \textbf{72.93} & \textbf{80.90} & \textbf{37.14} & 0.00 & \textbf{55.91} \\ \midrule
\multirow{2}{*}{Deepseek-R1-8B} & vanilla & 75.97 & 30.77 & 8.89 & 66.01 & 26.00 & 58.99 & \textbf{74.73} & 82.22 & \textbf{37.93} & \textbf{16.67} & \textbf{62.79} \\
 & instruct & \textbf{76.53} & \textbf{33.71} & \textbf{13.04} & \textbf{71.73} & \textbf{35.16} & \textbf{60.66} & 72.04 & \textbf{82.87} & 33.33 & 10.53 & 54.55 \\ \midrule
\multirow{2}{*}{Llama3.1-8B} & vanilla & 76.51 & 30.23 & 4.44 & 69.39 & 33.65 & 67.15 & 72.38 & \textbf{87.64} & \textbf{31.88} & \textbf{19.05} & 55.56 \\
 & instruct & \textbf{79.69} & \textbf{34.57} & \textbf{10.26} & \textbf{71.57} & \textbf{35.64} & \textbf{69.17} & \textbf{75.33} & 83.62 & 30.51 & 0.00 & \textbf{56.52} \\ \bottomrule
\end{tabular}
\end{adjustbox}
\caption{The F1 scores of IR-Tuning with vanilla inputs and instruction inputs on the ArgRevision (Evidence and Reasoning) and ITERATER corpora. The bold numbers are the best results in each setting. }

\label{tab:erevise-iterater-comparision}
\end{table*}

To investigate the efficiency of IR-Tuning, we plot the losses of fine-tuning Llama3.1-8B on the training sets of ArgRevision and ITERATER corpora in Figure~\ref{fig:training-loss-llama3.1-8b}. IR-Tuning exhibits fast convergence compared to other PEFT methods, of which LISA has surprisingly high fluctuations during the beginning steps, largely because it fine-tunes on random layers rather than the layers with high gradient norms. The similar patterns are observed on Mistral-7B and Deepseek-R1-8B in Figure~\ref{fig:training-loss-mistral-7b} and~\ref{fig:training-loss-deepseek-r1-8b} in the Appendix. These observations highlight the superiority of IR-Tuning for fast convergence. In addition, we plot its GPU memory allocations while fine-tuning on the ArgRevision corpus in Figure~\ref{fig:memory-cost} and the ITERATER corpus in Figure~\ref{fig:memory-cost-iterater} in the Appendix.  For a fair comparison, we set the batch size to one for both full fine-tuning and PEFT. The histogram shows that IR-Tuning with LoRA uses the smallest memory across different LLMs, compared to full-layer fine-tuning with and without LoRA, which confirms its memory efficiency.

\subsection{The Effectiveness of Instruction Tuning} 
We evaluate IR-Tuning performance on specific intentions with and without instruction tuning for each corpus. For ArgRevision, Table~\ref{tab:erevise-iterater-comparision} shows IR-Tuning is worse than RoBERTa in predicting \textit{Irrelevant} and mostly worse for \textit{Relevant} and \textit{Repeated} intentions, all from evidence revision. This might be because LLMs struggle to learn evidence revision with limited data; in contrast, LLMs perform well in predicting reasoning revision. Although Llama3.1-8B is generally better than Mistral-7B in several NLP tasks, it does not hold true on argument revisions using vanilla input data, which might be because Llama3.1-8B does not learn revisions well without proper instructions. However, IR-Tuning can improve Mistral-7B without using instruction tuning. For ITERATER, LLMs are generally better than the RoBERTa baseline except for \textit{Style}, which has limited annotations (e.g., 128 \textit{Style} labels in Table~\ref{table:iterator-corpus-stats1} in the Appendix). While instruction tuning is generally helpful on the ArgRevision corpus, except for Mistral-7B, it sometimes downgrades the performance on the ITERATER corpus for Deepseek-R1-8B and Llama3.1-8B. This might be because ArgRevision has substantial revisions that add and delete entire sentences (e.g., rows \#1 and \#2 in Table~\ref{table:data-example-argrevision} in the Appendix), thus IR-Tuning needs contextualized instructions to help predict intentions. In contrast, the revisions in ITERATER are minor (see Table~\ref{table:data-example-iterater} in the Appendix), which rely less on the instructions to learn intentions. These observations suggest that contextualized instructions are not always helpful for LLMs to learn revisions, which answers \textbf{RQ3}.

\begin{table}[tb]
\centering
\small
\begin{adjustbox}{width=0.98\columnwidth}
\begin{tabular}{cc|cc|cc}
\toprule
\multirow{2}{*}{Models} & \multirow{2}{*}{Sizes} & \multicolumn{2}{c|}{ArgRevision} & \multicolumn{2}{c}{ITERATER} \\ \cmidrule{3-6} 
& & F1-Score & AUPRC & F1-Score & AUPRC \\ \midrule
Phi-2 & 2.7B & 39.89 & 44.88 & 42.77 & 47.22 \\ 
Mistral & 7B & 47.20 & 49.16 & 49.38 & 54.22 \\ 
Deepseek-R1 & 8B & 48.47 & 50.14 & 50.66 & 54.46 \\ 
Llama3.1 & 8B& 50.15 & 52.69 & 49.20 & 56.42 \\ 
Llama2& 13B & \textbf{52.49} & \textbf{53.11} & \textbf{52.19} & \textbf{57.45} \\ \bottomrule
\end{tabular}
\end{adjustbox}
\caption{The performance of IR-Tuning with LoRA PEFT for different sizes of LLMs across two corpora.}
\label{tab:model_size}
\end{table}

\begin{table}[tb]
\centering
\small
\begin{adjustbox}{width=0.86\columnwidth}
\begin{tabular}{c|cc|cc}
\toprule
\multirow{2}{*}{Adapters} & \multicolumn{2}{c|}{ArgRevision} & \multicolumn{2}{c}{ITERATER} \\ \cmidrule{2-5} 
 & F1-Score & AUPRC & F1-Score & AUPRC \\ \midrule
Botteneck & 50.21 & \textbf{54.38} & 47.97 & 56.00 \\
LoRA & 50.15 & 52.69 & 49.20 & 56.42 \\
DoRA & \textbf{52.66} & 51.47 & \textbf{50.94} & \textbf{58.64} \\ \bottomrule
\end{tabular}
\end{adjustbox}
\caption{The performance of IR-Tuning on Llama3.1-8B with different PEFT adapters across two corpora.}
\label{tab:adapter_selection}
\end{table}

\subsection{Generalizability and Robustness}
To validate the generalizability of IR-Tuning, we evaluate its performance using LLMs with varying parameter sizes, ranging from small to large. Table~\ref{tab:model_size} shows that F1 scores and AUPRC generally improve as the size of the model increases, which implies that larger LLMs might work better since larger models can capture complex revision patterns and leverage richer contextual information. In addition, we implement IR-Tuning using different PEFT adapters, i.e., classic Bottleneck~\cite{pmlr-v97-houlsby19a} and advanced DoRA~\cite{liu-etal-dora-2024}. Table~\ref{tab:adapter_selection} shows that DoRA generally performs better than LoRA on two corpora, except for AUPRC in ArgRevision. This suggests that IR-Tuning benefits from DoRA’s ability to decompose the weight magnitude and direction within LLM layers~\cite{liu-etal-dora-2024}.

Furthermore, we compare our gradient-based importance scoring metric with the magnitude-based method regarding parameter weights~\cite{ma-etal-llm-pruner-2023} and the similarity-based method that measures the hidden state before and after LLM layers~\cite{chen2025streamlining}, as shown in Table~\ref{tab:importance_score_metrics}. Although the similarity metric appears to be advantageous on ArgRevision, it is more computationally expensive than magnitude and gradient-based metrics, as it necessitates additional computations right before and after each LLM layer. In addition, the gradient-based metric has a higher AUPRC on the ITERATER data, suggesting it's more reliable for unbalanced data, while the magnitude-based metric has generally accurate predictions on ITERATER. These observations suggest that performance varies across corpora; however, gradient-based methods are informative as they measure whether LLM layers are actively engaged during fine-tuning.  

\begin{figure}[tb]
    \centering
    \includegraphics[width=1\columnwidth]{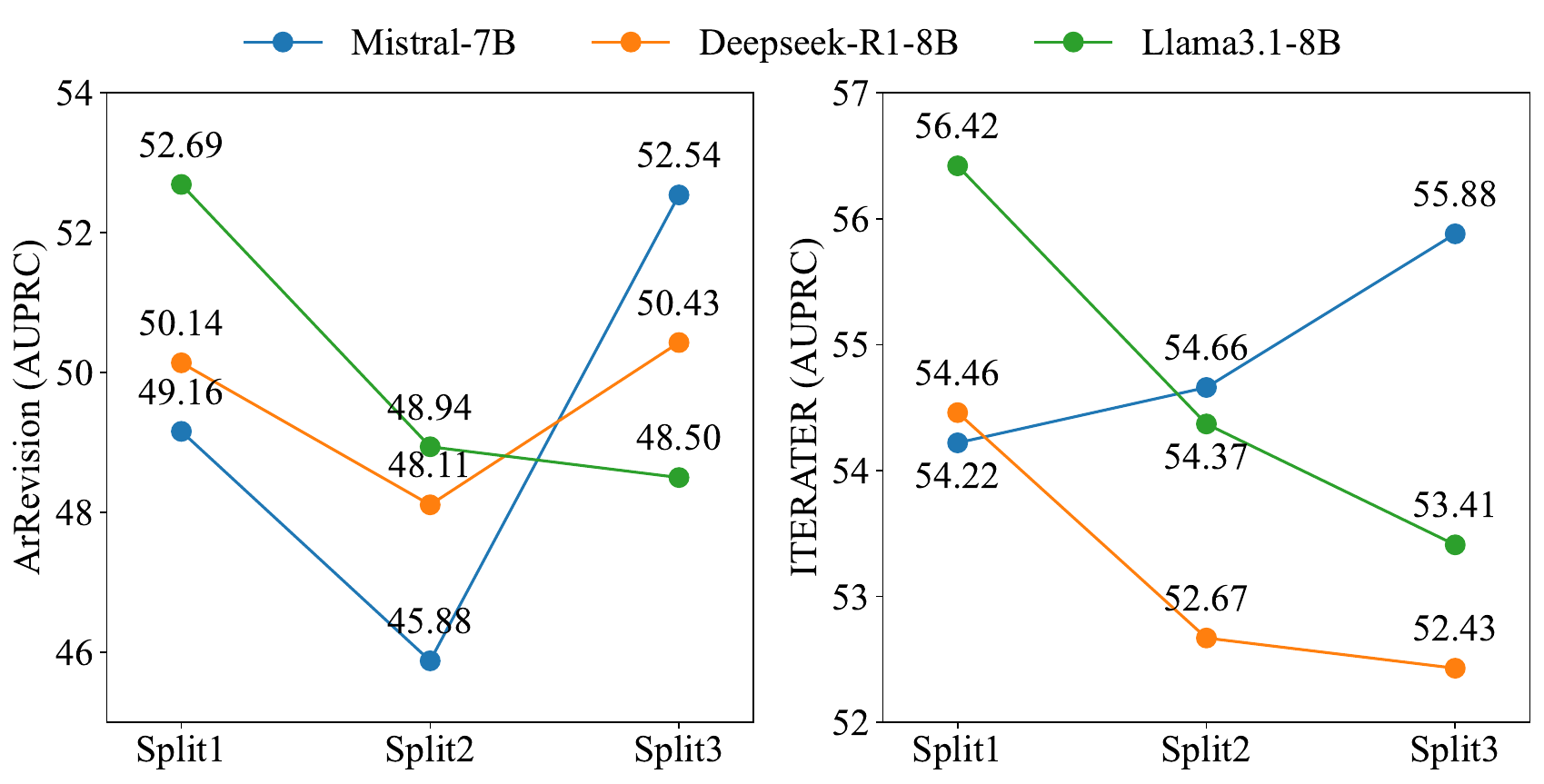}
    \caption{The performance (AUPRC) of IR-Tuning with LoRA PEFT and different splits across two corpora.}
    \label{fig:split-compare}
\end{figure}

Moreover, we study the parameter sensitivity of the layer-selection algorithm in Figure~\ref{fig:split-compare}. In the ArgRevision corpus, the performance of Llama3.1-8B decreases as adding more splits, which could be due to down-sampling issues, as shown in Figure~\ref{fig:over-down-sampling}. Mistral-7B and Deepseek-R1-8B downgrade performance in two splits but improve in three. In the ITERATER corpus, performance typically drops when adding more splits, except for Mistral-7B. These observations suggest that the layer selection depends on both data and models, where one split is generally good for Llama-based models, and three splits work best for Mistral-based models.

\begin{table}[tb]
\centering
\small
\begin{adjustbox}{width=0.92\columnwidth}
\begin{tabular}{c|cc|cc}
\toprule
\multirow{2}{*}{Model} & \multicolumn{2}{c|}{ArgRevision} & \multicolumn{2}{c}{ITERATER} \\ \cmidrule{2-5} 
 & F1-Score & AUPRC & F1-Score & AUPRC \\ \midrule
Similarity & \textbf{52.36} & \textbf{53.53} & 49.23 & 53.24 \\
Magnitude & 48.42 & 48.72 & \textbf{52.57} & 53.26 \\
Gradient & 50.15 & 52.69 & 49.20 & \textbf{56.42} \\ \bottomrule
\end{tabular}
\end{adjustbox}
\caption{The performance of IR-Tuning on Llama3.1-8B with LoRA PEFT and different importance scoring metrics across two corpora.}
\label{tab:importance_score_metrics}
\vspace{-.5em}
\end{table}

\section{Conclusion}
We present a layer-wise PEFT framework named IR-Tuning, which dynamically selects a subset of LLM layers for fine-tuning while freezing the others. 
We demonstrate that IR-Tuning is effective across multiple revision corpora and LLMs, and achieves fast fine-tuning convergence and consumes low GPU memory. Although IR-Tuning is primarily used for text revision in this study, it can potentially be used for other NLP tasks. 
In future work, we will evaluate its performance on more revision tasks to generalize our findings.

\section*{Ethics Statement}
We use our previously developed automated writing evaluation system~\cite{liu2025ereviserfwritingevaluationassessing} to collect the ArgRevision corpus under standard protocols approved by an Institutional Review Board (IRB). Expert annotators, familiar with both the system and the writing tasks, were employed to label revision intentions. The data collection and annotation processes do not raise any ethical concerns. Furthermore, our research on text revision, particularly for argument essay revisions, is critically needed for automated student writing evaluation, which benefits both the education and NLP communities.

\section*{Limitations}

Despite the effectiveness of our proposed method, several limitations are noted. First, the integration of IR-Tuning relies on the gradient norm on every LLM layer to hypothetically measure the importance of the layers. However, more data-driven and task-specific metrics, e.g., evaluation loss, could be leveraged to score layer importance. Nevertheless, IR-Tuning does not introduce significant computational overhead compared to standard PEFT, based on our pilot studies. We will further investigate its efficiency in future work. Also, we empirically use one split for selecting LLM layers in our current settings. We will explore automated methods for splitting layers in future work. Second, computational limitation constrains our evaluation to relatively lightweight models. While larger LLMs, e.g., Llama3.3-70B, offer enhanced capabilities for contextual understanding, we are unable to incorporate them in this study. Our evaluations are based on small annotated corpora, which have limited generalizability to cover diverse revisions in real-world scenarios. Third, we do not investigate other PEFT techniques, e.g., prompt-tuning. Thus, a deeper exploration of different PEFT methods, instruction designs, and diverse LLMs may achieve more robust task-specific fine-tuning strategies.

\section*{Acknowledgments}

This research was supported by National Science Foundation Award \#2202347 and its REU and NAIRR Pilot Demonstration Project supplements, as well as through CloudBank allocations. The opinions expressed are those of the authors and do not represent the views of the institutes. The authors would like to thank the anonymous reviewers and the Pitt PETAL groups for their valuable feedback on this work.

\bibliography{anthology,custom}

\appendix

\section{Appendix}
\label{sec:appendix}

\begin{table*}[!htb]\small
\centering
\begin{adjustbox}{width=1.85\columnwidth}
\begin{tabular}
{>{\centering\arraybackslash}m{0.01\textwidth}m{0.22\textwidth}m{0.22\textwidth}>{\centering\arraybackslash}m{0.08\textwidth}>{\centering\arraybackslash}m{0.08\textwidth}>{\centering\arraybackslash}m{0.08\textwidth}}	
\toprule
{ID} & {Sentence in Original Essay}                                                                                                                                                             & {Sentence in Revised Essay}                                                                                                                                                                                                                                & {Revision Behavior} & {Revision Type} & {Revision Intention} \\ \midrule
1           & Sauri has been struggling with poverty.                                                                                                                                                      &                                                                                                                                                                                                                                                                & Delete                     & Evidence               & Irrelevant                  \\ \midrule
2           &                                                                                                                                                                                              & The author did convince me that,"winning the fight against poverty is achievable in our life time"the author tells us that this is achievable in paragraph 3.in paragraph 3 it says," we are halfway to 2035, and the world is capable of meeting these goals. & Add                        & Evidence               & Relevant                    \\ \midrule
3           & ...                                                                                                                                                                                          & ...                                                                                                                                                                                                                                                            & ...                        & ...                    & ...                         \\ \midrule
4           &                                                                                                                                                                                              & I can infer they want to get Sauri out of poverty before 2035.                                                                                                                                                                                                 & Add                        & Reasoning              & Other                  \\ \midrule
5           & According to paragraph 3 it says,"the plan is to get people out of poverty, assure them access to health care and help them stabilize the economy and quality of life in their communities." & According to paragraph 3 it says,"the plan is to get people out of poverty, assure them access to health care and help them stabilize the economy and quality of life in their communities."                                                                   & N/A                        & N/A                    & N/A                         \\ \midrule
6           & This shows the Sauri needs as much help as they can to get out of poverty.                                                                                                                   & This shows the Sauri needs as much help as they can to get out of poverty.                                                                                                                                                                                     & N/A                        & N/A                    & N/A                         \\ \midrule
7           &                                                                                                                                                                                              & Also it shows that people are willing to help Sauri.                                                                                                                                                                                                           & Add                        & Reasoning              & LCE                         \\ \midrule
8           & ...                                                                                                                                                                                          & ...                                                                                                                                                                                                                                                            & ...                        & ...                    & ...                         \\ \midrule
9           & In 2035 I can infer that they will be financially stable and will have better housing.                                                                                                       & In 2035 I can infer that they will be financially stable and will have better housing.                                                                                                                                                                         & N/A                        & N/A                    & N/A                         \\ \midrule           10 & ...                                                                                                                                                                                          & ...                                                                                                                                                                                                                                                            & ...                        & ...                    & ...                         \\ \midrule
11          &                                                                                                                                                                                              & Local leaders take it from there."                                                                                                                     & Add                        & Evidence               & Relevant                    \\ \midrule       12    &                                                                                                                                                                                           & This shows how they provided Sauri with the necessities and they can see action taking place and see improvements.                                                                                                                                                                                                                                                            & Add                        & Reasoning                    & LCE                         \\ \bottomrule
\end{tabular}
\end{adjustbox}
\caption{The examples of revision intention annotations for an essay in the ArgRevision corpus.}
\label{table:data-example-argrevision}
\end{table*}

\begin{table*}[t]\small
\centering
\begin{adjustbox}{width=1.3\columnwidth}
\begin{tabular}{c|cccccc|c}
\toprule
       & Clarity & Fluency & Coherence & Style & Meaning & Others & Total \\ \midrule
Add    & 94      & 208     & 37        & 2     & 342             & 6      & 689   \\
Delete & 282     & 111     & 180       & 20    & 11              & 11     & 615   \\
Modify & 1,225    & 623     & 176       & 106   & 543             & 41     & 2,714  \\ \midrule
Total  & 1,601    & 942     & 393       & 128   & 896             & 58     & 4,018  \\ \bottomrule
\end{tabular}
\end{adjustbox}
\caption{The statistics of revisions in the ITERATER corpus.}
\label{table:iterator-corpus-stats1}
\end{table*}

\begin{table*}[!htb]\small
\centering
\begin{adjustbox}{width=1.85\columnwidth}
\begin{tabular}
{>{\centering\arraybackslash}m{0.01\textwidth}m{0.3\textwidth}m{0.3\textwidth}>{\centering\arraybackslash}m{0.12\textwidth}}	
\toprule
ID & Sentence in Original Article                                                                                                                                                                                                                                                                                                                                                                                                                       & Sentence in Revised Article                                                                                                                                                                                                                                                                                                               & Revision Intention \\ \midrule
1  & In this work, we point out the inability to infer behavioral conclusions from probing results, and offer an alternative   method which is focused on how the information is being used, rather than on   what information is encoded.                                                                                                                                                                                                              & In   this work, we point out the inability to infer behavioral conclusions from   probing results, and offer an alternative method which focuses on how the   information is being used, rather than on what information is encoded.                                                                                                      & Style              \\ \midrule

2 &  In this work, we point out the inability to infer behavioral conclusions from probing results , and offer an alternative method which focuses on how the information is being used, rather than on what information is encoded. &  In this work, we point out the inability to infer behavioral conclusions from probing results and offer an alternative method which focuses on how the information is being used, rather than on what information is encoded. & Fluency \\ \midrule

3 &  In this work, we point out the inability to infer behavioral conclusions from probing results , and offer an alternative method which focuses on how the information is being used, rather than on what information is encoded. &  In this work, we point out the inability to infer behavioral conclusions from probing results , and offer an alternative method that focuses on how the information is being used, rather than on what information is encoded. & Clarity \\ \midrule

4  & A growing body of work makes use of probing in order to investigate the working of neural models, often   considered black boxes.                                                                                                                                                                                                                                                                                                                  & A growing body of work makes use of probing to investigate the working of   neural models, often considered black boxes.                                                                                                                                                                                                                  & Coherence          \\ \midrule

5  & We explore representations from different model families (BERT, RoBERTa, GPT-2 , etc.) and find evidence for emergence of linguistic manifold across layer depth (e.g., manifolds for part-of-speech and combinatory categorical grammar tags). We further observe that different encoding schemes used to obtain the representations lead to differences in whether these linguistic manifolds emerge in earlier or later layers of the network . & We explore representations from different model families (BERT, RoBERTa, GPT-2 , etc.) and find evidence for emergence of linguistic manifold across layer depth (e.g., manifolds for part-of-speech tags), especially in ambiguous data (i.e, words with multiple part-of-speech tags, or part-of-speech classes including many words) . & Meaning-changed   \\ \bottomrule
\end{tabular}
\end{adjustbox}
\caption{The examples of revision intention annotations in the ITERATER corpus.}
\label{table:data-example-iterater}
\end{table*}

\begin{table*}[t]\small
\centering
\begin{adjustbox}{width=1.2\columnwidth}
\begin{tabular}{c|ccc|ccc}
\toprule
\multirow{2}{*}{} & \multicolumn{3}{c|}{Evidence Revision} & \multicolumn{3}{c}{Reasoning Revision} \\ \cmidrule{2-7} 
                  & Relevant   & Irrelevant   & Repeated   & LCE       & not LCE    & Commentary    \\ \midrule
Train             & 2,029      & 425          & 216        & 1,200     & 358        & 506           \\
Val               & 240        & 50           & 29         & 148       & 46         & 79            \\
Test              & 240        & 50           & 31         & 144       & 56         & 71            \\ \midrule
Total             & 2,509      & 525          & 276        & 1,492     & 460        & 656           \\ \bottomrule
\end{tabular}
\end{adjustbox}
\caption{\label{table:sentence-stats-split}
The statistics of revision intentions in the training, validation, and test sets of the ArgRevision corpus.
}
\end{table*}

\begin{table*}[t]\small
\centering
\begin{adjustbox}{width=1\columnwidth}
\begin{tabular}{c|ccccc}
\toprule
           & Clarity & Fluency & Coherence & Style & Meaning  \\ \midrule
Train       & 1,258    & 739     & 311       & 100   & 807         \\
Val  & 157     & 115     & 46        & 13    & 54     \\
Test       & 186     & 88      & 36        & 15    & 35          \\ \midrule
Total    & 1,601   & 942     & 393       & 128   & 896          \\\bottomrule
\end{tabular}

\end{adjustbox}
\caption{The statistics of revision intentions in the training, validation, and test sets of the ITERATER corpus.}
\label{table:iterator-corpus-stats-split}
\end{table*}

\begin{figure*}[t]
    \centering
    \includegraphics[width=1\linewidth]{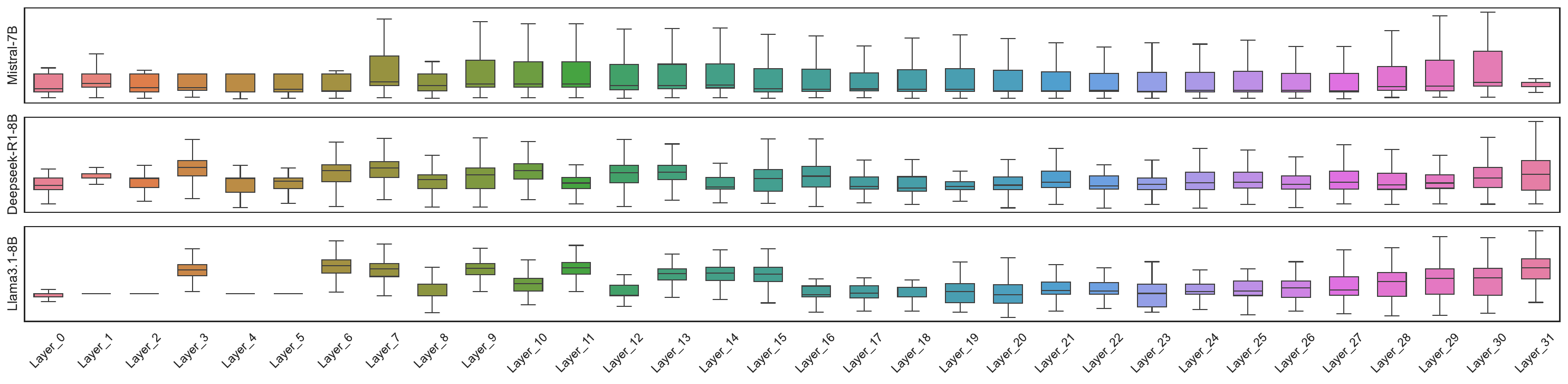}
    \caption{The importance scores (gradient norms) vary across different LLM layers on the ITERATER corpus. The high variances indicate that the gradients have changed significantly, suggesting the layers are actively learning from the data. The layers with low variances mean they have been frozen most of the time.}
    \label{fig:gradient-change-iterater}
\end{figure*}

\begin{figure*}[t]
    \centering
    \includegraphics[width=1.25\columnwidth]{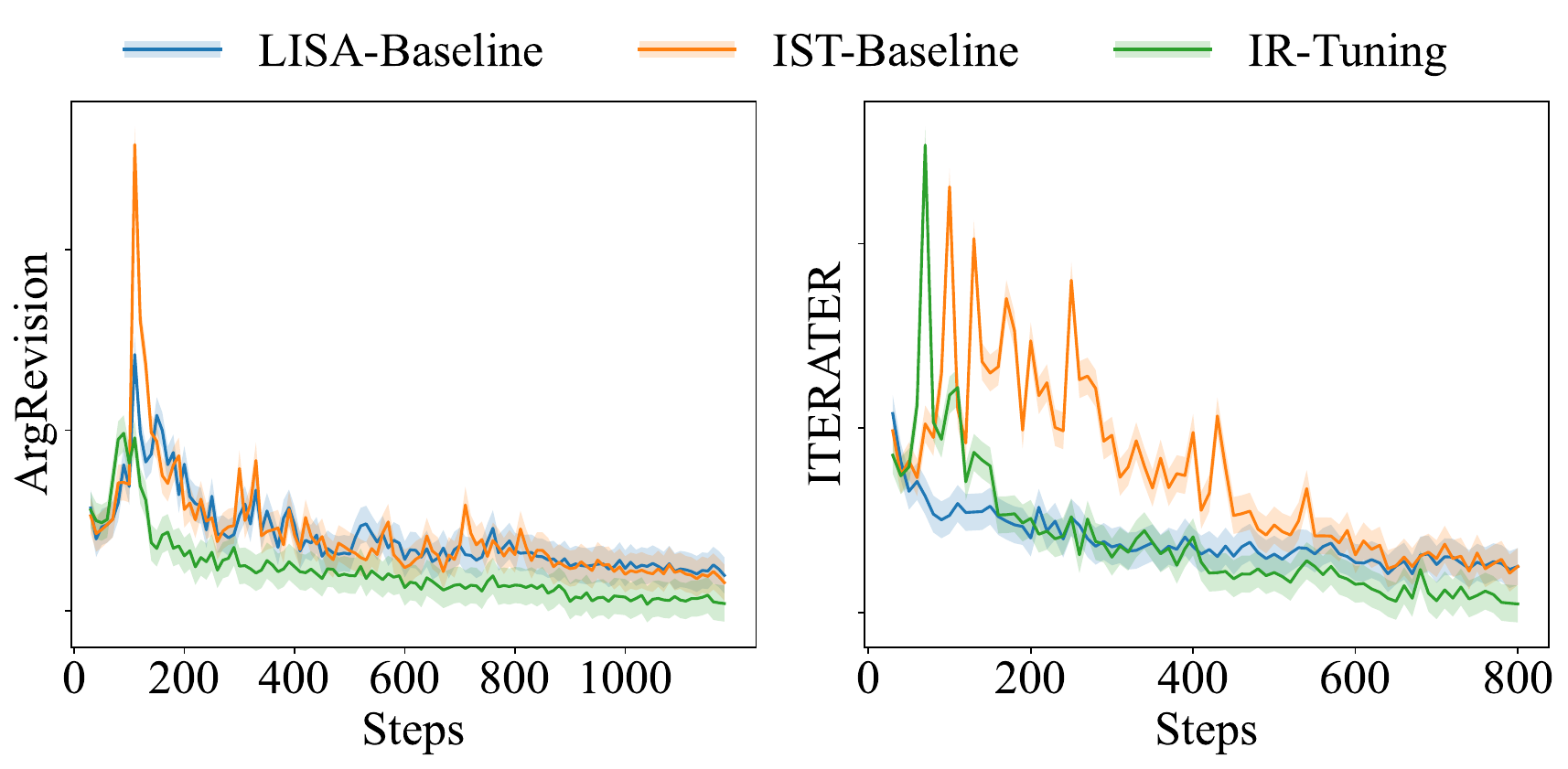}
    \caption{The Mistral-7B fine-tuning losses on the ArgRevision and ITERATER training sets, using IR-Tuning and baseline PEFT methods.}
    \label{fig:training-loss-mistral-7b}
\end{figure*}

\begin{figure*}[t]
    \centering
    \includegraphics[width=1.25\columnwidth]{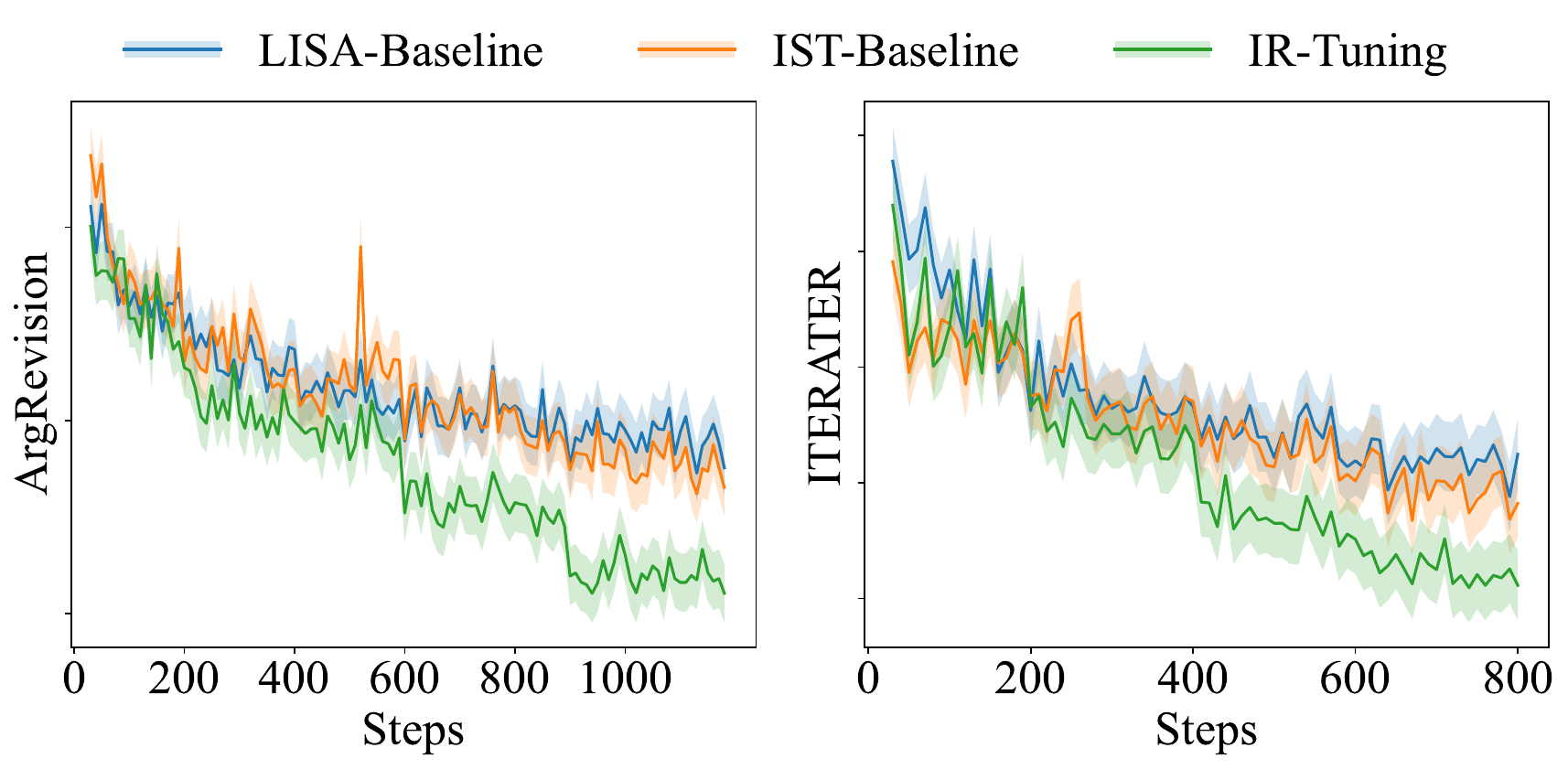}
    \caption{The Deepseek-R1-8B fine-tuning losses on the ArgRevision and ITERATER training sets, using IR-Tuning and baseline PEFT methods.}
    \label{fig:training-loss-deepseek-r1-8b}
\end{figure*}

\begin{figure*}[t]
    \centering
    \includegraphics[width=1.3\columnwidth]{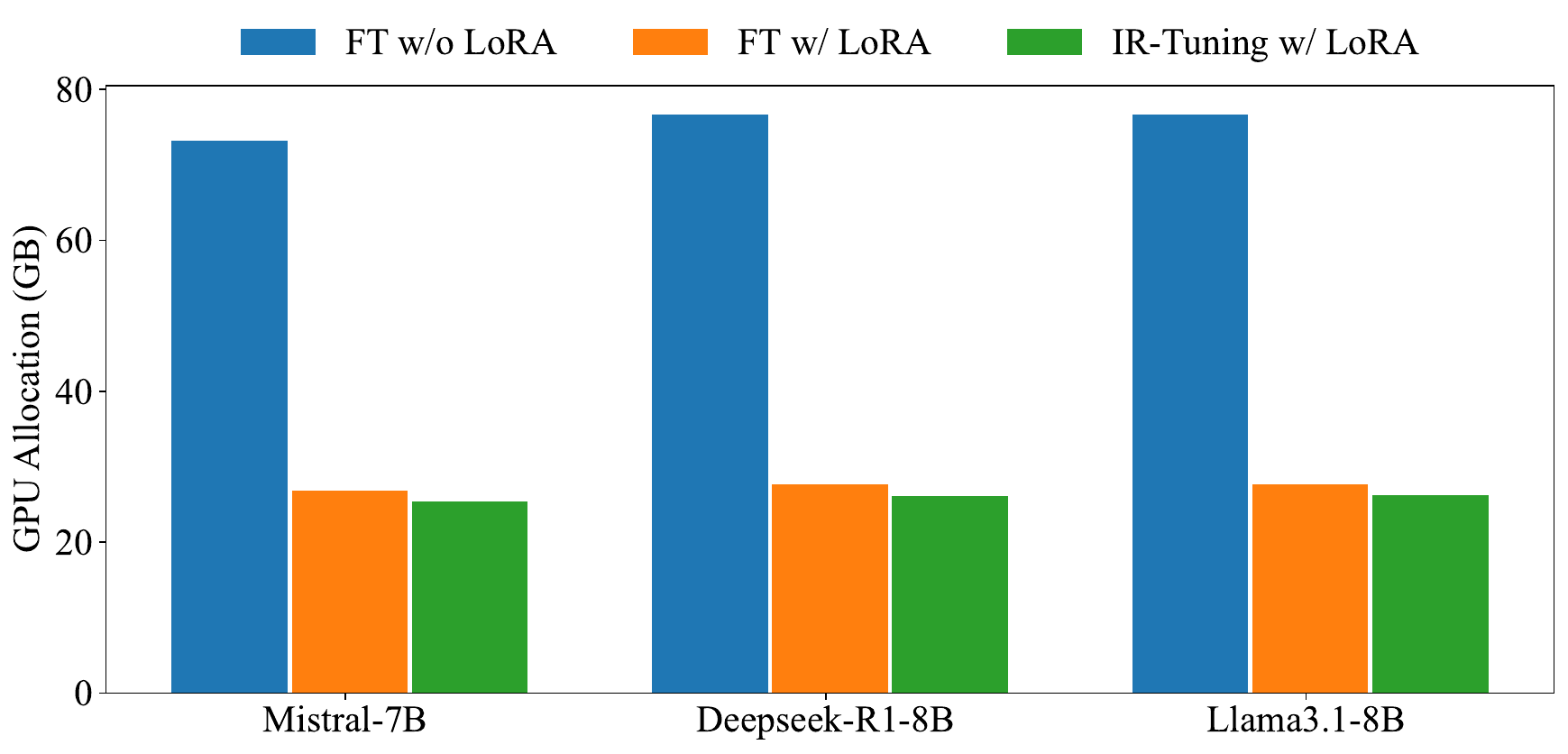}
    \caption{The GPU memory allocations for the Full-Finetuning (FT) with and without PEFT (LoRA), and IR-Tuning with PEFT (LoRA) on the ITERATER corpus.}
    \label{fig:memory-cost-iterater}
\end{figure*}

\end{document}